\def\year{2021}\relax
\documentclass[letterpaper]{article} 
\usepackage{aaai21}  
\usepackage{times}  
\usepackage{helvet} 
\usepackage{courier}  
\usepackage[hyphens]{url}  
\usepackage{graphicx} 
\usepackage{natbib}  
\usepackage{caption} 
\usepackage{amsthm,amsmath,amssymb}
\usepackage{mathrsfs}
\usepackage{multirow}
\usepackage{booktabs}
\usepackage{url}
\title{ROSITA: Refined BERT cOmpreSsion with InTegrAted techniques}
\author{

    Yuanxin Liu\textsuperscript{\rm 1,2} 
    Zheng Lin\textsuperscript{\rm 1} \thanks{Zheng Lin is the corresponding author of this paper.}
    Fengcheng Yuan\textsuperscript{\rm 3}\thanks{Contribution made at IIE, CAS.}
    \\
}
\affiliations{

    \textsuperscript{\rm 1}Institute of Information Engineering, Chinese Academy of Sciences, Beijing, China\\
    \textsuperscript{\rm 2}School of Cyber Security, University of Chinese Academy of Sciences, Beijing, China \\
    \textsuperscript{\rm 3}Meituan Inc.

    \{liuyuanxin,linzheng\}@iie.ac.cn, yuanfengcheng@meituan.com

}

\begin{document}

\maketitle

\begin{abstract}
Pre-trained language models of the BERT family have defined the state-of-the-arts in a wide range of NLP tasks. However, the performance of BERT-based models is mainly driven by the enormous amount of parameters, which hinders their application to resource-limited scenarios. Faced with this problem, recent studies have been attempting to compress BERT into a small-scale model. However, most previous work primarily focuses on a single kind of compression technique, and few attention has been paid to the combination of different methods. When BERT is compressed with integrated techniques, a critical question is how to design the entire compression framework to obtain the optimal performance. In response to this question, we integrate three kinds of compression methods (weight pruning, low-rank factorization and knowledge distillation (KD)) and explore a range of designs concerning model architecture, KD strategy, pruning frequency and learning rate schedule. We find that a careful choice of the designs is crucial to the performance of the compressed model. Based on the empirical findings, our best compressed model, dubbed \textbf{R}efined BERT c\textbf{O}mpre\textbf{S}sion with \textbf{I}n\textbf{T}egr\textbf{A}ted techniques (ROSITA), is $7.5 \times$ smaller than BERT while maintains $98.5\%$ of the performance on five tasks of the GLUE benchmark, outperforming the previous BERT compression methods with similar parameter budget. The code is available at \url{https://github.com/llyx97/Rosita}.
\end{abstract}

\section{Introduction}
Pre-trained language models (PLMs) have become a popular research topic in recent years, enjoying great breakthroughs in a wide range of NLP tasks. One of the major driving forces behind the PLMs is the huge number of model parameters. Typical PLMs like BERT \cite{DevlinCLT19}, GPT \cite{GPT} and XLNet \cite{XLNet} have hundreds of millions of parameters. However, the overparamterized PLMs also suffer from expensive computational overhead, making it intractable to deploy them to resource-limited scenarios.

In response to the above problem, efforts have been paid to reduce the size of PLMs. \citet{CompressingBERT} showed that $30-40 \%$ of BERT's weights can be pruned without affecting the performance on downstream tasks. TinyBERT \cite{TinyBERT} and MobileBERT \cite{MobileBERT} use knowledge distillation (KD) \cite{Hinton} to transfer the power of BERT into compact models. They compress BERT to approximately 15M parameters with only a marginal drop in performance. Recently, further progress has been made for both the pruning \cite{MovementPruning} and KD methods \cite{BERT-EMD,MiniLM}.

In spite of these successes, most existing BERT compression studies primarily focus on a single technique, while few attention has been paid to the system designs when multiple compression techniques work as a whole. Since some compression techniques are naturally complementary to each other, a combination of them can integrate their respective advantages. Therefore, it is important to study BERT compression in the context of integrated techniques.

When multiple techniques are integrated to compress BERT, a critical question is how to design the entire compression framework. To this end, we combine weight pruning, low-rank factorization and KD, and investigate a range of design choices that concern model architecture, KD strategy, pruning frequency and learning rate schedule. Through extensive experiments we find that: 1) Different dimensions of the BERT structure (e.g., layers, neurons and attention heads) should be compressed at different ratios. Generally, a deep architecture produces better results than a wide architecture under the same overall compression ratio. 2) It is beneficial to divide KD into multiple stages. In each stage, we fully train the student and let it serve as the teacher in the next stage. In this way, we can provide better supervision for the students in the subsequent stages and alleviate the negative effect caused by the structural discrepancy between teacher and student. 3) In integrated BERT compression, the pruning frequency (i.e. the speed at which BERT is compressed) has a double-edged effect on the final performance and the learning rate schedule plays a role in this process.

\begin{figure*}[t]
\centering
\includegraphics[width=0.7\textwidth]{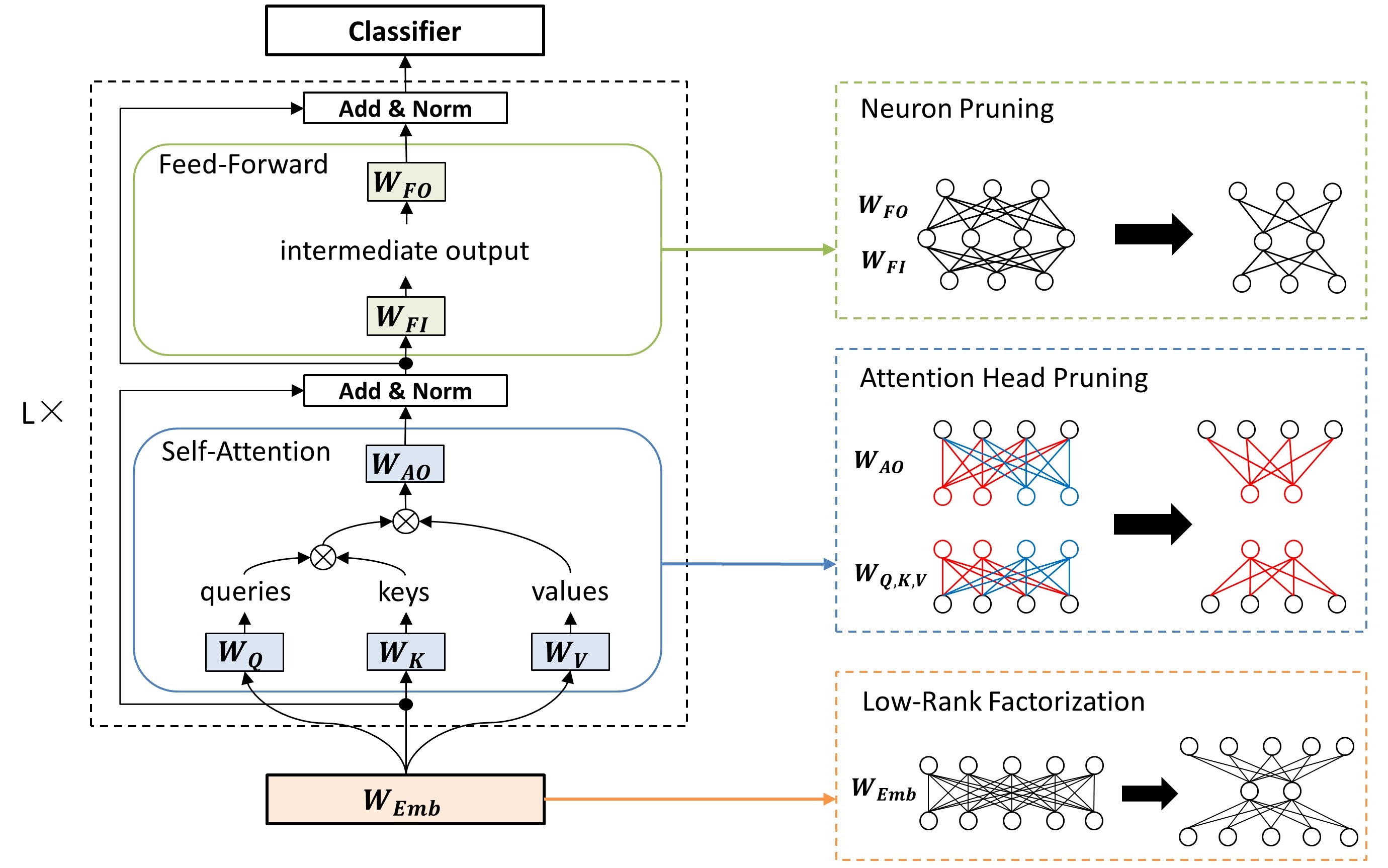}
\caption{Left: The architecture of BERT. The colored boxes are weight matrices that can be compressed. Right: Illustration of neuron pruning (upper), attention head pruning (middle) and factorization of the word embedding matrix (lower).}
\label{fig:bert}
\end{figure*}

Based on these findings, we show that BERT compression with integrated techniques can achieve promising results. Given the same parameter budget, our best compressed model outperforms the state-of-the-art BERT compression methods on five tasks of the GLUE benchmark.

\section{Background: BERT Architecture}
The left part of Figure \ref{fig:bert} shows BERT's architecture, which is comprised of an embedding layer, $L$ identical Transformer \cite{Transformer} layers, and a task-specific classifier.

Each Transformer layer consists of two sub-layers: the self-attention layer and the feed-forward network (FFN). The self-attention layer has $H$ paralleled attention heads, each head is formulated as a self-attention operation:
\begin{equation}
    \begin{aligned}&\operatorname{Self-Att}_{i}(X)=\operatorname{Att}(X W_{Qi}, X W_{Ki}, X W_{Vi}) \\ 
    &\operatorname{Att}(Q,K,V)=\operatorname{softmax}\left(\frac{Q K^{\top}}{\sqrt{d_{k}}}\right) V\end{aligned}
\end{equation}
where $X \in \mathbb{R}^{|X| \times d_{X}}$ is the layer input, $|X|$ is the sequence length and $d_{X}$ is the hidden dimension of BERT. $W_{Qi}$, $W_{Ki}$, $W_{Vi} \in \mathbb{R}^{d_{X} \times \frac{d_{X}}{H}}$ are weight matrices that transform $X$ into the attention spaces.

The outputs of different heads are then concatenated and linearly projected to obtain the output of self-attention:
\begin{equation}
\resizebox{1.0\hsize}{!}{$
\text { MultiHead }(X)=\left[\operatorname{Self-Att}_{1}(X); \ldots; \operatorname{Self-Att}_{H}(X)\right] W_{AO}
$}
\end{equation}
where $W_{AO} \in \mathbb{R}^{d_{X} \times d_{X}}$ is the output matrix of the self-attention layer and $[;]$ denotes the concatenation operation. In practice, the weight matrices $W_{Qi}$, $W_{Ki}$, $W_{Vi}$ of different heads will be combined into three large matrices $W_{Q}$, $W_{K}$, $W_{V}$, so as to execute the $H$ heads in parallel.

The FFN is composed of two successive linear transformations with a ReLU activation in between:
\begin{equation}
\mathrm{FFN}(X)=\max \left(0, X W_{FI}+b_{FI}\right) W_{FO}+b_{FO}
\end{equation}
where $W_{FI} \in \mathbb{R}^{d_{X} \times d_{I}}$, $W_{FO} \in \mathbb{R}^{d_{I} \times d_{X}}$ and $b_{FI,FO}$ are the learnable weights and biases, respectively. $d_{I}$ is the dimension of the intermediate outputs. Following each self-attention layer and FFN, there are a series of dropout, residual connection and layer normalization. 

Finally, the output of the Transformer module is fed to the task-specific classifier and the entire model is trained with cross-entropy loss:
\begin{equation}
\mathcal{L}_{cross}
=-y^{*} \cdot \log \left(\operatorname{softmax}\left(z\right)\right)
\end{equation}
where $y^{*}$ is the one-hot vector of ground-truth label and $z$ is the predicted logits.

\section{Integrated Techniques for BERT Compression}
In this paper, we consider three kinds of compression techniques: weight pruning, low-rank factorization and knowledge distillation. These techniques have different advantages that can complement each other in BERT compression. In knowledge distillation \cite{Hinton}, a small student model is trained to mimic the behaviours of a large teacher model. Initializing the student with certain parts of the teacher provides a better start point for KD, which necessitates the technique of weight pruning to drop unimportant weights. Low-rank factorization plays an important role in compressing the embedding matrix, which will be explained later. We do not consider weight sharing since it can only reduce the model parameters but not the computational complexity.

Specifically, we compress BERT from four kinds of structural dimensions: Transformer layer, the intermediate outputs of the FFN ($W_{FI,FO}$), attention heads ($W_{Q}$, $W_{K}$, $W_{V}$), and the token embedding matrix ($W_{Emb}$). In terms of the layer dimension, we simply keep the first $L^{'}$ layers for compression. The other three dimensions will now be discussed.

\subsection{Weight Pruning}
Weight pruning is used to compress the FFN intermediate outputs and the attention heads. We adopt structured weight pruning, where the less important neurons or entire attention heads are pruned.

To determine the importance of neurons, we first measure the importance of each single weight, and then calculate the neuron importance by adding up the importance of its connected weights. Following \citet{Pavlo}, we compute the weight importance score using a metric based on first-order Taylor expansion:
\begin{equation}
\label{equ:taylor}
S_{W_{ij}} = \mathbb{E}_{x \sim \mathcal{D}} \left|\frac{\partial \mathcal{L}(x)}{\partial W_{ij}} W_{ij}\right|
\end{equation}
where $W_{ij}$ is the weight in the $i^{th}$ row and $j^{th}$ column of matrix $W$. $x$ is a data example from the dataset $\mathcal{D}$. Theoretically, this metric approximates the change of loss function $\mathcal{L}$ when removing a particular weight.

For the FFN, the weights from both $W_{FI}$ and $W_{FO}$ that are connected to an intermediate neuron will contribute to the neuron importance. The importance of the $i^{th}$ attention head is computed as the sum of importance of the neurons in $\operatorname{Self-Att}_{i}(X)$. For these neurons, we only consider the weights in $W_{AO}$ to compute the importance. Based on the importance scores, we can eliminate the less important neurons or attention heads given an expected compression ratio.

Practically, there are two typical ways of weight pruning: pruning to the final compression ratio in one step and iterative pruning, in which the expectation in Equation \ref{equ:taylor} is computed differently. For \textbf{one-step pruning}, we average the first-order Taylor expansion scores over the training set. For \textbf{iterative pruning}, we accumulate the scores over the training steps between pruning operations.

\subsection{Low-rank Factorization}
Since there are residual connection after each sub-layer, we keep the hidden dimension $d_{X}$ unchanged. However, the embedding matrix, whose output dimension equals to $d_{X}$, takes up a large proportion of the model parameters. Inspired by \citet{ALBERT,Ziheng}, we decompose the embedding matrix into two smaller matrices using Singular Value Decomposition (SVD):
\begin{equation}
W_{Emb}= U \Sigma V = \sum_{i=1}^{r} \sigma_{i} \times\left(U_{i} \times V_{i}\right)
\end{equation}
where $U \in \mathbb{R}^{|\mathcal{V}| \times r}$ and $V \in \mathbb{R}^{r \times d_{X}}$ are the decomposed matrices, and $|\mathcal{V}|$ is vocabulary size. $\Sigma=\operatorname{diag}\left(\sigma_{1}, \sigma_{2}, \ldots, \sigma_{r}\right)$ is a diagonal matrix consisting of the singular values $\sigma_{i}$, where $r \leq \min (|\mathcal{V}|, d_{X})$ is the matrix rank. $U_{i}$ and $V_{i}$ are the $i^{th}$ column of $U$ and $i^{th}$ row of $V$ respectively. 

The importance of each $U_{i}$ and $V_{i}$ is measured according to the corresponding $\sigma_{i}$. By dropping the less effective columns or rows, we can reduce the size of $U$ and $V$.

\subsection{Knowledge Distillation}
We select two kinds of BERT knowledge to transfer, namely the predictions and the hidden states. The predictions are learned through a soft cross-entropy loss function:
\begin{equation}
\mathcal{L}_{pred }=-\operatorname{softmax}\left(z^{T}\right) \cdot \log \left(\operatorname{softmax}\left(z^{S}\right)\right)
\end{equation}
where $z^{T}$ and $z^{S}$ are the predicted logits of teacher and student, respectively. The teacher's hidden states are learned by minimizing the Mean Squared Error (MSE):
\begin{equation}
\mathcal{L}_{h i d d e n}=\sum_{l=0}^{L^{'}} \operatorname{MSE}\left(X^{T}_{g(l)}, X^{S}_{l}\right)
\end{equation}
where $L^{'}$ is the student layer number, and the $0^{th}$ layer is the embedding layer. $X^{S}_{l}$ is the outputs of the $l^{th}$ student layer.  $g(l)$ is the mapping function to select the teacher layers to learn. When $\bmod (L, L^{'})= 0$, we select the teacher layers evenly, i.e., $g(l)=l \times \frac{L}{L^{'}}$. Otherwise, we follow \citet{DynaBERT} to drop the teacher layers that satisfy $\bmod (l+1, \frac{L}{L^{'}})= 0$, and knowledge from the reminding $L^{'}$ layers will be distilled into the student model.

\begin{table}[t]
\small
\centering
\begin{tabular}{@{}l c c c c @{}}
\toprule
Dataset    &Train   &Train(aug)  &Dev       &Test \\\midrule
CoLA       &8,551   &213,080    &1,043       &1,063    \\
SST-2      &67,349  &1.1M      &873         &1,821\\
QNLI       &104,743  &4.2M      &5,463       &5,463\\
QQP        &363,870 &7.6M      &40,431      &390,965\\
MNLI-m     &392,702 &8.0M      &9,815       &9,796\\
MNLI-mm    &392,702 &8.0M      &9,832       &9,847\\

\bottomrule
\end{tabular}
\caption{Dataset statistics.}
\label{tab:dataset}
\end{table}

\section{General Experimental Setup}
\subsection{Datasets}
The General Language Understanding Evaluation (GLUE) benchmark \cite{GLUE} is a collection of diverse tasks, including natural language inference and sentiment analysis, etc. We evaluate the compression methods on five tasks: CoLA, SST-2, QNLI, QQP and MNLI. Based on the original training sets, we utilize the data augmentation algorithm proposed by \citet{TinyBERT} to construct augmented training sets for knowledge distillation. The data statistics are shown in Table \ref{tab:dataset}.

\subsection{Evaluation}
In our exploratory study, we evaluate the performance on the development sets. The evaluation metrics are Matthews correlation (mcc) for CoLA, Accuracy for SST-2, QNLI, QQP and MNLI-m/mm. For comparison with existing BERT compression methods, we also report results on the held-out test sets, which are obtained from the GLUE test server \footnote{https://gluebenchmark.com/leaderboard}. 

\begin{table*}[t]
\small
\centering
\begin{tabular}{@{}l l c c c c c c c c@{}}
\toprule
& Architecture   &Size  & CoLA & SST-2 & QNLI & QQP & MNLI-m & MNLI-mm & Avg\\ \midrule

& $\mathrm{BERT}_{\mathrm{BASE}}$ ($H=12,L=12,d_{I}=3072,r=768$) 
&109M   &59.1   &93.0   &91.5   &90.9   &84.6  &84.6   &84.0 \\
\midrule

&A: $\underline{H=9},L=3,d_{I}=768,r=128$         
&13.9M   &14.0    &86.0   &83.5   &87.3   &74.5 &74.9    &70.0\\
&B: $\underline{H=3},L=3,d_{I}=1088,r=192$       
&13.8M   &+0.1    &-0.3    &-0.6   &-0.4    &-1.1 &-0.5    &-0.5\\
\midrule

&C: $H=2,\underline{L=8},d_{I}=512,r=128$       
&14.5M    &\bf20.2   &\bf87.6   &\bf83.8   &\bf88.3  &\bf77.7   &\bf78.0   &\bf72.6\\
&D: $H=4,\underline{L=3},d_{I}=1024,r=192$        
&14.1M    &-8.0   &-2.6   &-0.8   &-1.2   &-3.0  &-3.6   &-3.2\\
\midrule

&E: $H=3,L=3,\underline{d_{I}=1664},r=128$        
&14.5M   &13.6   &85.5   &81.8   &86.6  &72.7  &73.8    &69.0\\
&F: $H=4,L=4,\underline{d_{I}=256},r=256$   
&13.7M   &-0.5   &+0.1   &+1.6   &+1.1  &+2.3   &+1.0   &+0.9\\
\midrule

&G: $H=2,L=2,d_{I}=576,\underline{r=320}$   
&13.6M   &13.8   &83.8   &62.3   &79.6  &60.7  &60.4   &60.1\\
&H: $H=4,L=4,d_{I}=1216,\underline{r=64}$   
&13.6M   &-1.3   &+0.4   &+20.3  &+7.7  &+13.0  &+13.9   &+9.0\\
\midrule

&I: $H=3,L=4,d_{I}=768,r=192$   
&14.1M   &14.4   &86.3   &83.4   &87.6  &73.9  &74.7   &70.1\\
\bottomrule
\end{tabular}
\caption{Dev set results of one-step pruning with different architectures (averaged over 3 runs). The compressed models are divided into five parts. Each of the first four parts focus on a certain structural dimension of interest. The underlined dimension is larger in size in the first row of each part, and smaller in the second row. The change in scores (against the first row) is shown in the second rows. \textbf{Lower} score change means the dimension is \textbf{more} important. The last part evenly compresses the four dimensions. The best scores are shown in bold font.}
\label{tab:architecture}
\end{table*}

\subsection{Implementation Details}
In this paper, we focus on the compression of $\mathrm{BERT}_{\mathrm{BASE}}$. We fine-tune $\mathrm{BERT}_{\mathrm{BASE}}$ and the compressed models on the five GLUE tasks using the HuggingFace transformers library \footnote{https://github.com/huggingface/transformers}. We modify the code released by \citet{TinyBERT}\footnote{https://github.com/huawei-noah/Pretrained-Language-Model/tree/master/TinyBERT} to perform knowledge distillation, using the fine-tuned $\mathrm{BERT}_{\mathrm{BASE}}$ as the original teacher. The models are trained with Adam optimizer \cite{Adam}. The training hyperparameters are tuned separately for each task. We select the value of learning rate from $\{1e-5, 2e-5, 3e-5, 5e-5\}$ and the value of batch size from $\{32, 64\}$. The range of the number of epoch varies across different settings. Due to space limitation, please refer to the code link for detailed hyperparameter settings.

\section{The Effect of Model Architecture}
As we discussed in the Background, BERT has four major structural dimensions, which play different roles. Therefore, it is intuitively that the four dimensions could be compressed at different ratios, which gives rise to different model architectures. In this section, we will investigate the effect of architecture in BERT compression.

\subsection{Experimental Setup}
We fix the size of the compressed model to 13.5M$\sim$14.5M, and compare the performance of different architectures. We consider two settings: 1) \textbf{One-step pruning}: Compressing BERT in one step and conduct fine-tuning in the original training set using $L_{cross}$. 2) \textbf{One-step pruning + one-stage KD}: Training the compressed model with KD objectives ($\mathrm{BERT}_{\mathrm{BASE}}$ as the teacher) on the augmented training sets.

\subsection{Results}
Table \ref{tab:architecture} presents the results of one-step pruning. The compressed models are divided into five parts. Each of the first four parts contains two architectures, where the first one has a larger size in the underlined dimension and the second one is smaller in this dimension. To focus on the effect of a particular dimension, the compression ratio of the rest three dimensions are kept roughly equivalent for each architecture. The last part is an architecture where the size of the four dimensions are roughly the same. As we can see, there is a perceivable variation in performance. For the attention head and layer, shrinking the size has a negative impact, while smaller intermediate outputs and embedding matrix rank result in better performance. This suggests that $d_{I}$ and $r$ can be compressed to a larger extent. Especially, the deepest architecture ($H$=2,$L$=8,$d_{I}$=512,$r$=128) consistently outperforms the other architectures, including the evenly compressed one, which indicates that different dimensions should be compressed at different ratios. 

We further validate the above finding when the KD technique is included. Compared with Table \ref{tab:architecture}, the results in Table \ref{tab:pruning+kd} exhibit a clear improvement, which demonstrates the complementary effect of KD with weight pruning and factorization. In the KD setting, the deeper Architecture C again outperforms the wider Architecture I, with an obvious improvement of 8 points in CoLA. Therefore, we can conclude that a deep and narrow model is a better choice over a wide and shallow model in BERT compression. Based on this conclusion, we use Architecture C in later experiments.

\begin{table}[t]
\small
\centering
\begin{tabular}{@{}l c c c c @{}}
\toprule
Model     &CoLA  &SST-2  &MNLI-m/mm  &Avg \\\midrule
Architecture I
&47.0      &92.3       &81.2/81.0   &75.4\\

Architecture C      
&\bf55.0   &\bf92.8    &\bf83.2/83.5   &\bf78.6\\

\bottomrule
\end{tabular}
\caption{Dev set results of one-step pruning + KD. The notations C and I are inherited from Table \ref{tab:architecture}.}
\label{tab:pruning+kd}
\end{table}

\section{Multi-Stage Knowledge Distillation}
In vanilla KD, the teacher model is trained on the original training set with $\mathcal{L}_{cross}$. However, the power of BERT may be limited when the number of training samples is inadequate. Moreover, when we compress the layer dimension, the discrepancy of layer number between teacher and student may affect hidden state knowledge distillation.

In response to the first problem, we distill the knowledge from the original BERT teacher to a model of the same size. With the abundant augmented data, this student can learn the downstream tasks better and is less likely to overfit. Then, the well-learned student serves as the teacher for the next KD stage. To better distill the hidden state knowledge to the final student, we introduce an intermediate student which has the same layer number as the final student. This intermediate student can achieve desirable results by only learning the teacher's predictions, which bypasses the difficulties in hidden state KD caused by layer number gap. 

\begin{figure}[t]
\centering
\includegraphics[width=1.0\linewidth]{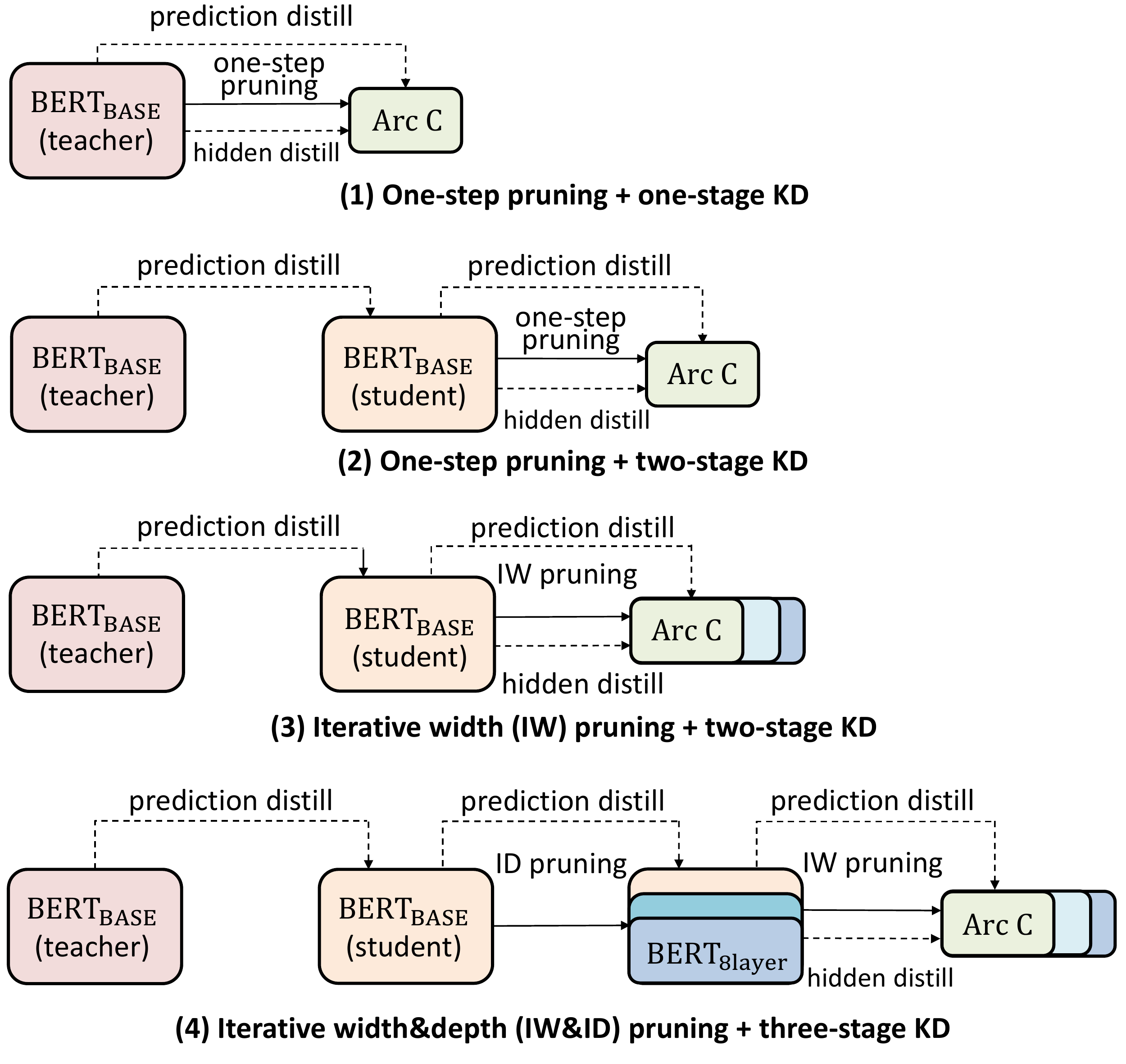} 
\caption{Illustration of different pruning and KD settings. Arc C denotes Architecture C.}
\label{fig:multi-stage}
\end{figure}

\subsection{Experimental Setup}
To examine the effectiveness of the above solutions, we explore a range of multi-stage KD strategies, which are illustrated in Figure \ref{fig:multi-stage}:
\begin{itemize}
\item \textbf{One-step pruning+one-stage KD}: Compressing BERT in one step and training the compressed model with KD.

\item \textbf{One-step pruning+two-stage KD}: All structural dimensions are pruned in one step like one-step pruning + one-stage KD. For KD, the knowledge in $\mathrm{BERT}_{\mathrm{BASE}}$ (teacher) is first transferred to $\mathrm{BERT}_{\mathrm{BASE}}$ (student), which has the same size as $\mathrm{BERT}_{\mathrm{BASE}}$ (teacher). Then, Architecture C distills the knowledge from $\mathrm{BERT}_{\mathrm{BASE}}$ (student).

\item \textbf{Iterative width pruning+two-stage KD}: The attention heads, intermediate outputs and embedding matrix (collectively referred to as the width dimensions) are iteratively pruned in the second stage of KD.

\item \textbf{Iterative width\&depth pruning+three-stage KD}: The layer is iteratively pruned in the second KD stage with $\mathrm{BERT}_{\mathrm{BASE}}$ (student) as teacher, which gives rise to $\mathrm{BERT}_{\mathrm{8layer}}$. The other three dimensions are pruned in the third KD stage, where the teacher is $\mathrm{BERT}_{\mathrm{8layer}}$. 
\end{itemize}
For all settings, KD is performed on the augmented training set. Hidden knowledge is only distilled in the last stage, and the teacher predictions are learned throughout all KD stages. 

\begin{table}[t]
\small
\centering
\begin{tabular}{@{}l c c c c @{}}
\toprule
Model   &CoLA  &SST-2  &MNLI-m/mm  &Avg \\\midrule

$\mathrm{BERT}_{\mathrm{BASE}}$ (teacher)
&59.1      &\bf93.0    &84.6/84.6    &80.3\\

$\mathrm{BERT}_{\mathrm{BASE}}$ (student) 
&\bf62.9   &\bf93.0    &\bf85.3/85.3 &\bf81.6\\
\midrule

Architecture C          \\
\quad one-stage  
&55.0   &92.8    &83.2/83.5   &78.6\\
\quad two-stage      
&\bf55.3   &\bf93.1    &\bf83.4/83.6 &\bf78.9\\

\bottomrule
\end{tabular}
\caption{The effect of multi-stage KD when BERT is compressed in one step. Dev set results are reported.}
\label{tab:two-stage-one-step}
\end{table}
\begin{table}[t]
\small
\centering
\begin{tabular}{@{}l c c c c @{}}
\toprule
Model  &CoLA  &SST-2  &MNLI-m/mm  &Avg \\\midrule

$\mathrm{BERT}_{\mathrm{8layer}}$  w/ ID
&61.4   &93.1    &84.6/85.1  &81.1\\
\midrule

Architecture C         \\
\quad two-stage IW
&55.9   &93.1    &83.4/\bf83.7    &79.0         \\

\quad three-stage IW\&ID
&\bf57.2   &\bf93.3    &\textbf{83.5}/83.6    &\bf79.4\\

\bottomrule
\end{tabular}
\caption{The effect of multi-stage KD with iterative pruning. IW and ID denote iterative width pruning and iterative depth pruning, respectively. Dev set results are reported.}
\label{tab:two-stage-iterative}
\end{table}

\subsection{Results}
To examine whether multi-stage KD can alleviate the negative effect of training data scarcity, we first compare the performance of $\mathrm{BERT}_{\mathrm{BASE}}$ (student) with $\mathrm{BERT}_{\mathrm{BASE}}$ (teacher). As shown in Table \ref{tab:two-stage-one-step}, the student $\mathrm{BERT}_{\mathrm{BASE}}$ outperforms its teacher, which suggests that BERT is undertrained on the original task-specific training sets. We attribute this improvement to KD and data augmentation, with which $\mathrm{BERT}_{\mathrm{BASE}}$ (student) can better capture the data distribution and is less likely to overfit. Thanks to the improved capability of $\mathrm{BERT}_{\mathrm{BASE}}$ (student), the final student is provided with better supervision. As a result, the final performance of two-stage KD outstrips one-stage KD.

To bridge the gap of layer number in hidden state knowledge distillation, we introduce another KD stage. Table \ref{tab:two-stage-iterative} presents the results of two-stage and three-stage KD under the iterative pruning setting. Comparing with the results in Table \ref{tab:two-stage-one-step}, we can derive three observations. First, although $\mathrm{BERT}_{\mathrm{8layer}}$ has fewer layers than $\mathrm{BERT}_{\mathrm{BASE}}$, it can still improve over $\mathrm{BERT}_{\mathrm{BASE}}$. This can be explained by the better supervision provided by $\mathrm{BERT}_{\mathrm{BASE}}$ (student) and the access to augmented training data. Second, iteratively pruning the width dimensions produces slightly better results than one-step pruning. The advantage of iterative pruning has also been recognized in the literature of model compression. In the next section, we will delve into the interaction of this pruning strategy with learning rate schedule. Lastly, using $\mathrm{BERT}_{\mathrm{8layer}}$ as the third stage teacher circumvents the difficulty in distilling hidden state knowledge, and further outperforms the iterative width pruning strategy.

\section{The Effect of Pruning Frequency and Learning Rate Schedule}
The pruning frequency is a key hyperparamter in iterative pruning. When KD and iterative pruning are performed simultaneously, the pruning frequency and learning rate schedule exert a joint impact on the performance. In this section, we study this impact through the combination of different learning rate schedules and pruning frequencies.

\begin{figure}[t]
\centering
\includegraphics[width=0.9\linewidth]{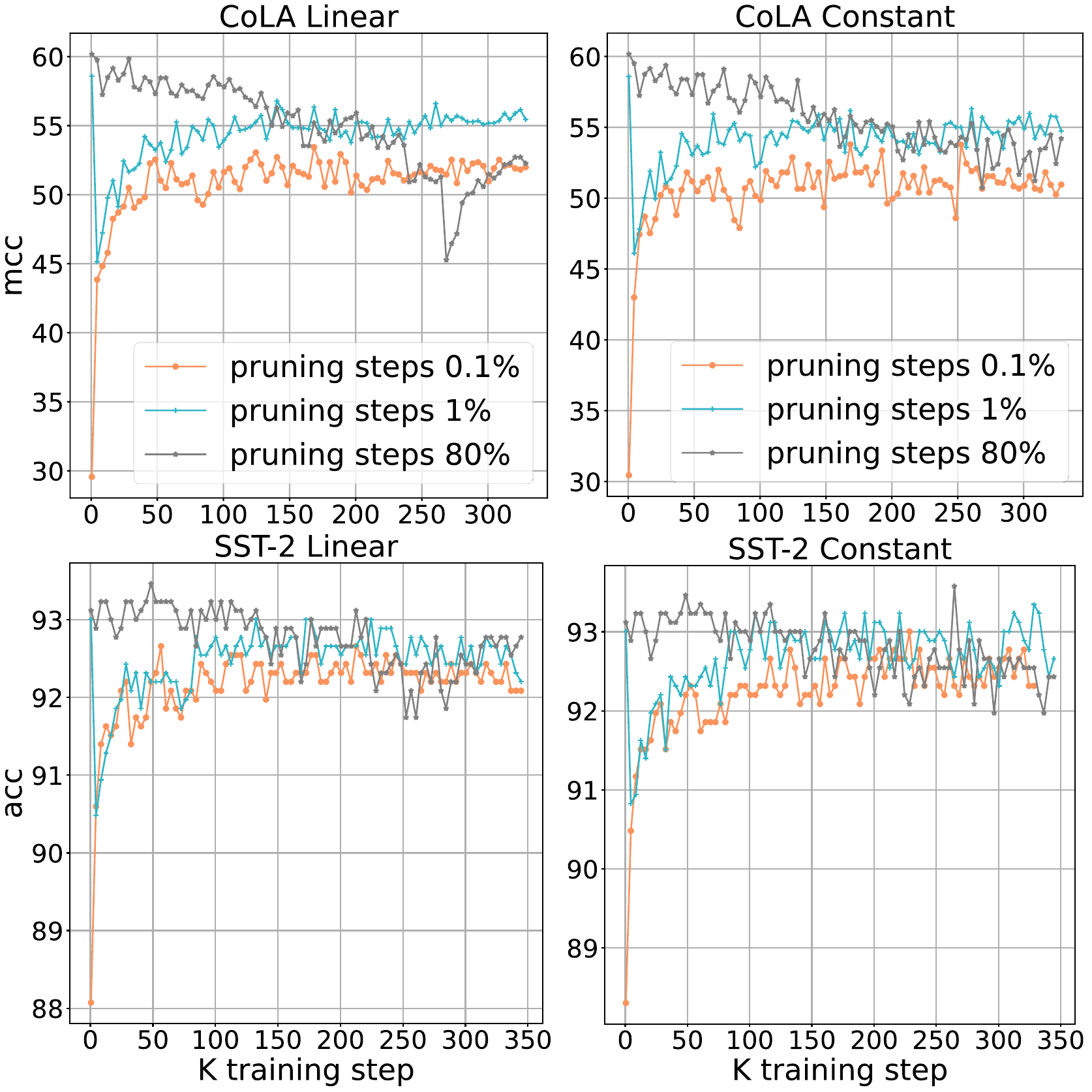} 
\caption{Mcc/acc curves in the third KD stage.}
\label{fig:pruning-speed}
\end{figure}

\subsection{Experimental Setup}
We fix the total number of training steps and assign a certain percentage of training steps for iterative pruning (referred to as pruning steps). A small percentage means the pruning frequency is high and vice versa. During the pruning steps, the student model will be pruned for a fixed number of times to the target size. Every time a certain proportion of the model is pruned. The numbers of training steps between every two pruning operations are equal. For example, the initial student is $\mathrm{BERT}_{\mathrm{8layer}}$, the final student is Architecture C, and the number of pruning steps is 1,000. We prune the model 10 times every 100 training steps. Every time 1 attention head, 256 intermediate neurons and 64 embedding matrix ranks will be pruned. For iterative depth pruning, we prune one layer each time. We consider two kinds of learning rate schedules: constant schedule and linear decaying schedule. The experiments of this section are performed under the last KD setting as described in the prior section. 

\subsection{Results}
We can see from Figure \ref{fig:pruning-speed} that the pruning frequency plays an important role in the performance. Compared with pruning rapidly (orange line) or slowly (grey line), an intermediate level of pruning frequency (blue line) performs the best. This is because the impact of pruning frequency is double-edged. On the one hand, a lower pruning frequency leads to a higher lower bound of performance (i.e., the performance at the end of iterative pruning), which provides a better start point for recovering. On the other hand, pruning too slowly leaves few training steps for recovering. We have explored a range of pruning step proportions and find the best result to be achieved at $1\% \sim 10\%$. For clear illustration, we only present 3 curves in Figure \ref{fig:pruning-speed}. 

The learning rate schedule also exerts its influence. When the learning rate decays during training, the speed for recovering turns slower every time after pruning. Therefore, when the proportion of pruning steps is set to $80\%$, the linear decaying schedule has a lower performance lower bound as compared with the constant learning rate. For CoLA, this impact is considerable and the remaining $20\%$ of training steps is not enough to recover the performance. In SST-2, although for linear schedule the performance drop during iterative pruning is greater than constant schedule, the performance is still kept at a recoverable level at $80\%$ of training.

\section{Comparison with Existing Methods}

\subsection{Baselines}
\begin{itemize}
\item \textbf{DynaBERT} \cite{DynaBERT} is based on task-specific KD and structured weight pruning. It compresses the attention head, intermediate outputs and Transformer layer, while the embedding matrix is left untouched.

\item \textbf{TinyBERT} \cite{TinyBERT} is a pure KD method that performs KD in both the pre-training stage and the task-specific fine-tuning stage. 

\item \textbf{MobileBERT} \cite{MobileBERT} combines task-agnostic KD and a bottleneck architecture, with an inverted-bottleneck teacher that is 2.7 $\times$ larger than $\mathrm{BERT}_{\mathrm{BASE}}$.

\item \textbf{MiniLM} \cite{MiniLM} is another task-agnostic KD method for BERT compression. In addition to distilling the key-query relationship in self-attention, it also learns the value-value relationship from the teacher.

\item \textbf{LadaBERT} \cite{LadaBERT} is an initial attempt to integrate weight pruning, low-rank factorization and KD.
\end{itemize}

It is worth noticing that some of the baselines differ from ROSITA in training or pruning setup. MobileBERT and MiniLM do not require task-specific KD, while ROSITA replaces pre-training KD with pruning and factorization. The KD in LadaBERT is done without data augmentation, while the weight pruning is unstructured. Although the settings are different and the methods are not strictly comparable, here we focus on the performance of the complete systems.

\begin{table*}[t]
\small
\centering

\begin{tabular}{@{}l l c c c c c c c c c c@{}}
\toprule
& Method   &Size  &\# of layers &Infer Time & CoLA & SST-2 & QNLI & QQP & MNLI-m & MNLI-mm & Avg\\ \midrule

\multirow{5}{*}{Dev}
& $\mathrm{BERT}_{\mathrm{BASE}}$*
&109M($\times1.0$)  &12 &-  &59.1  &93.0   &91.5   &90.9   &84.6  &84.6 &83.9\\
\cmidrule{2-12}

&DynaBERT        
&34M($\times3.2$)   &6  &-  &43.7    &92.0   &88.5   &\bf90.4   &82.0   &82.3  &79.8\\

&MiniLM         
&22M($\times5.0$)   &6  &-  &-  &91.2   &-   &-   &82.4  &-  &- \\

&TinyBERT      
&14.5M($\times7.5$)  &4  &-  &50.8   &92.7*   &89.1*   &90.1*  &82.8  &82.9     &81.4   \\

&MobileBERT      
&15.1M($\times7.2$)  &24  &-  &-   &91.6  &89.9   &-   &82.0   &-   &-\\\

&ROSITA
&14.5M($\times7.5$)  &8  &-  &\bf57.2   &\bf93.3   &\bf90.2   &90.3  &\bf83.5    &\bf83.6   &\bf83.0 \\
\midrule

\multirow{5}{*}{Test} 
& $\mathrm{BERT}_{\mathrm{BASE}}$
&109M($\times1.0$)  &12  &874s($\times1.0$)  &52.1   &93.5   &90.5   &71.2   &84.6  &83.4   &79.2\\
\cmidrule{2-12}

&LadaBERT 
&15M($\times7.3$)   &12  &-  &-   &89.9   &84.5   &69.4   &82.1  &81.8 &-\\

&TinyBERT      
&14.5M($\times7.5$)  &4  &117s($\times7.5$) &44.1   &92.6   &87.7   &\bf71.3  &82.5  &81.8     &76.7\\

&MobileBERT      
&15.1M($\times7.2$)  &24  &290s($\times3.0$) &46.7   &91.7   &\bf89.5   &68.9  &81.5   &81.6    &76.7\\

&ROSITA
&14.5M($\times7.5$)  &8  &207s($\times4.2$) &\bf47.4   &\bf93.3   &\bf89.5  &\bf71.3  &\bf83.7   &\bf82.9   &\bf78.0\\

\bottomrule
\end{tabular}
\caption{Comparison with state-of-the-art BERT compression methods. * means our implementation. The metric for QQP in the test set is F1. The inference time is measured on a single 24GB TITAN RTX GPU over the original MNLI training set (the batch size and maximum sequence length are set to 128).}
\label{tab:bert-compress}
\end{table*}

\subsection{Results}
As can be seen in Table \ref{tab:bert-compress}, DynaBERT achieves comparable performance with other methods with doubled size. This is primarily because it does not compress the embedding matrix, which poses a limit on the compression ratio. Compared with the methods with similar model size, ROSITA outperforms MobileBERT and TinyBERT that rely on KD, and LadaBERT that also integrates three compression techniques. This suggests that a careful design of the integrated compression framework is crucial. In general, ROSITA is $7.5 \times$ smaller than $\mathrm{BERT}_{\mathrm{BASE}}$ and maintains $98.9\%$ (dev) and $98.5\%$ (test) of the average performance. In addition to parameter count, the model depth has a considerable impact on the inference speed. ROSITA is slower than TinyBERT, which has only four layers. However, to compensate the negative effect caused by dropping too much layers, TinyBERT requires KD in both the pre-training and task-specific stages, which increases the training budget. Compared with MobileBERT, ROSITA is both faster and performs better.

\section{Related Work}
To facilitate the deployment of BERT to resource-constrained scenarios, recent studies have explored a variety of compression techniques. \citet{CompressingBERT} prune $30\% \sim 40\%$ of BERT's weight without affecting the performance on downstream tasks. \citet{Paul} observe similar effect in attention head pruning. \citet{Ziheng} apply SVD to the weight matrices of BERT in the fine-tuning stage, which reduces the number of parameters by $35\%$ and keeps nearly $99\%$ for the performance. ALBERT \cite{ALBERT} combines embedding matrix factorization and cross-layer weight sharing in the pre-training stage, the resulting 18M ALBERT is comparable with the 109M $\mathrm{BERT}_{\mathrm{BASE}}$. Movement Pruning \cite{MovementPruning} measures the weight importance based on their change in magnitude during fine-tuning, which achieves $95\%$ sparsity of the encoder weights. However, weight sharing and unstructured sparsity cannot be directly translated to inference speedup. Therefore, there is still room for improvement for the above methods.

To recover the drop in performance caused by parameter reduction, knowledge distillation \cite{Hinton} is introduced to the pre-training stage \cite{MobileBERT}, the task-specific training stage \cite{PKD,DistillBiLSTM,DynaBERT,Well-read}, or both stages \cite{DistillBERT,TinyBERT}. Recently, some progress has been made in the design of KD losses. MiniLM \cite{MiniLM} proposed to distill the value-value relationship in self-attention. BERT-EMD \cite{BERT-EMD} allows the student to learn from any teacher layers using the Earth Mover's Distance. As an alternative to KD, BERT-OF-THESEUS \cite{BERT-of-Theseus} randomly replaces the modules of a model with compact modules during training.

Different from the the researches that focus on a single compression technique, we study the design choices for an integrated BERT compression framework. Related work includes DynaBERT \cite{DynaBERT}, which combines task-specific KD with structured weight pruning. Movement Pruning also enhances its proposed pruning metric with KD. Similar to our work, LadaBERT \cite{LadaBERT} also combines weight pruning, matrix factorization and KD. However, the weights are pruned in an unstructured way. Concurrent to our work, \citet{TernaryBERT} combines KD and quantization. The 2-bit quantized TernaryBERT is comparable with the full-precision model, which also attest to the importance of integrating different compression techniques.

Our multi-stage KD strategy is related to the method proposed by \citet{Seyed}, which introduces a teacher assistant in multi-stage KD. However, they aim at bridging the capability gap between teacher and student. In contrast, our motivation is to provide better supervision for the final student, which requires an intermediate model with stronger capability than the teacher. Moreover, we bridge the gap in layer number, instead of the capability gap, in an attempt to facilitate hidden state knowledge distillation.

\section{Conclusion}
In this paper, we highlight the importance of using integrated techniques to compress BERT. Under an integrated compression framework, we investigate the effect of a range of designs concerning model architecture, KD strategy, pruning frequency and learning rate schedule. We show that a careful choice of the designs is crucial to the performance of the compressed model, which provides a reference for future studies on this topic. Based on the empirical findings, we build a novel BERT compression framework named ROSITA, which outperforms the previous state-of-the-art methods with similar parameter constraint.

\section{Acknowledgments}
This work was supported by National Natural Science Foundation of China (No. 61976207, No. 61906187).

\bibliography{aaai2021}

\begin{thebibliography}{26}
\providecommand{\natexlab}[1]{#1}
\providecommand{\url}[1]{\texttt{#1}}
\providecommand{\urlprefix}{URL }
\expandafter\ifx\csname urlstyle\endcsname\relax
  \providecommand{\doi}[1]{doi:\discretionary{}{}{}#1}\else
  \providecommand{\doi}{doi:\discretionary{}{}{}\begingroup
  \urlstyle{rm}\Url}\fi

\bibitem[{Devlin et~al.(2019)Devlin, Chang, Lee, and Toutanova}]{DevlinCLT19}
Devlin, J.; Chang, M.; Lee, K.; and Toutanova, K. 2019.
\newblock {BERT:} Pre-training of Deep Bidirectional Transformers for Language
  Understanding.
\newblock In \emph{{NAACL-HLT} {(1)}}, 4171--4186. Association for
  Computational Linguistics.

\bibitem[{Gordon, Duh, and Andrews(2020)}]{CompressingBERT}
Gordon, M.~A.; Duh, K.; and Andrews, N. 2020.
\newblock Compressing {BERT:} Studying the Effects of Weight Pruning on
  Transfer Learning.
\newblock In \emph{RepL4NLP@ACL}, 143--155. Association for Computational
  Linguistics.

\bibitem[{Hinton, Vinyals, and Dean(2015)}]{Hinton}
Hinton, G.~E.; Vinyals, O.; and Dean, J. 2015.
\newblock Distilling the Knowledge in a Neural Network.
\newblock \emph{CoRR} abs/1503.02531.

\bibitem[{Hou et~al.(2020)Hou, Huang, Shang, Jiang, Chen, and Liu}]{DynaBERT}
Hou, L.; Huang, Z.; Shang, L.; Jiang, X.; Chen, X.; and Liu, Q. 2020.
\newblock DynaBERT: Dynamic {BERT} with Adaptive Width and Depth.
\newblock In \emph{NeurIPS}.

\bibitem[{Jiao et~al.(2020)Jiao, Yin, Shang, Jiang, Chen, Li, Wang, and
  Liu}]{TinyBERT}
Jiao, X.; Yin, Y.; Shang, L.; Jiang, X.; Chen, X.; Li, L.; Wang, F.; and Liu,
  Q. 2020.
\newblock TinyBERT: Distilling {BERT} for Natural Language Understanding.
\newblock In \emph{{EMNLP} (Findings)}, 4163--4174. Association for
  Computational Linguistics.

\bibitem[{Kingma and Ba(2015)}]{Adam}
Kingma, D.~P.; and Ba, J. 2015.
\newblock Adam: {A} Method for Stochastic Optimization.
\newblock In \emph{{ICLR} (Poster)}.

\bibitem[{Lan et~al.(2020)Lan, Chen, Goodman, Gimpel, Sharma, and
  Soricut}]{ALBERT}
Lan, Z.; Chen, M.; Goodman, S.; Gimpel, K.; Sharma, P.; and Soricut, R. 2020.
\newblock {ALBERT:} {A} Lite {BERT} for Self-supervised Learning of Language
  Representations.
\newblock In \emph{{ICLR}}. OpenReview.net.

\bibitem[{Li et~al.(2020)Li, Liu, Zhao, Xu, Yang, and Jin}]{BERT-EMD}
Li, J.; Liu, X.; Zhao, H.; Xu, R.; Yang, M.; and Jin, Y. 2020.
\newblock {BERT-EMD:} Many-to-Many Layer Mapping for {BERT} Compression with
  Earth Mover's Distance.
\newblock In \emph{{EMNLP} {(1)}}, 3009--3018. Association for Computational
  Linguistics.

\bibitem[{Mao et~al.(2020)Mao, Wang, Wu, Zhang, Wang, Zhang, Yang, Tong, and
  Bai}]{LadaBERT}
Mao, Y.; Wang, Y.; Wu, C.; Zhang, C.; Wang, Y.; Zhang, Q.; Yang, Y.; Tong, Y.;
  and Bai, J. 2020.
\newblock LadaBERT: Lightweight Adaptation of {BERT} through Hybrid Model
  Compression.
\newblock In \emph{{COLING}}, 3225--3234. International Committee on
  Computational Linguistics.

\bibitem[{Michel, Levy, and Neubig(2019)}]{Paul}
Michel, P.; Levy, O.; and Neubig, G. 2019.
\newblock Are Sixteen Heads Really Better than One?
\newblock In \emph{NeurIPS}, 14014--14024.

\bibitem[{Mirzadeh et~al.(2020)Mirzadeh, Farajtabar, Li, Levine, Matsukawa, and
  Ghasemzadeh}]{Seyed}
Mirzadeh, S.; Farajtabar, M.; Li, A.; Levine, N.; Matsukawa, A.; and
  Ghasemzadeh, H. 2020.
\newblock Improved Knowledge Distillation via Teacher Assistant.
\newblock In \emph{{AAAI}}, 5191--5198. {AAAI} Press.

\bibitem[{Molchanov et~al.(2017)Molchanov, Tyree, Karras, Aila, and
  Kautz}]{Pavlo}
Molchanov, P.; Tyree, S.; Karras, T.; Aila, T.; and Kautz, J. 2017.
\newblock Pruning Convolutional Neural Networks for Resource Efficient
  Inference.
\newblock In \emph{{ICLR} (Poster)}. OpenReview.net.

\bibitem[{Radford et~al.(2018)Radford, Narasimhan, Salimans, and
  Sutskever}]{GPT}
Radford, A.; Narasimhan, K.; Salimans, T.; and Sutskever, I. 2018.
\newblock Improving language understanding with unsupervised learning.
\newblock In \emph{{Technical report, OpenAI}}.

\bibitem[{Sanh et~al.(2019)Sanh, Debut, Chaumond, and Wolf}]{DistillBERT}
Sanh, V.; Debut, L.; Chaumond, J.; and Wolf, T. 2019.
\newblock DistilBERT, a distilled version of {BERT:} smaller, faster, cheaper
  and lighter.
\newblock \emph{CoRR} abs/1910.01108.

\bibitem[{Sanh, Wolf, and Rush(2020)}]{MovementPruning}
Sanh, V.; Wolf, T.; and Rush, A.~M. 2020.
\newblock Movement Pruning: Adaptive Sparsity by Fine-Tuning.
\newblock In \emph{NeurIPS}.

\bibitem[{Sun et~al.(2019)Sun, Cheng, Gan, and Liu}]{PKD}
Sun, S.; Cheng, Y.; Gan, Z.; and Liu, J. 2019.
\newblock Patient Knowledge Distillation for {BERT} Model Compression.
\newblock In \emph{{EMNLP/IJCNLP} {(1)}}, 4322--4331. Association for
  Computational Linguistics.

\bibitem[{Sun et~al.(2020)Sun, Yu, Song, Liu, Yang, and Zhou}]{MobileBERT}
Sun, Z.; Yu, H.; Song, X.; Liu, R.; Yang, Y.; and Zhou, D. 2020.
\newblock MobileBERT: a Compact Task-Agnostic {BERT} for Resource-Limited
  Devices.
\newblock In \emph{{ACL}}, 2158--2170. Association for Computational
  Linguistics.

\bibitem[{Tang et~al.(2019)Tang, Lu, Liu, Mou, Vechtomova, and
  Lin}]{DistillBiLSTM}
Tang, R.; Lu, Y.; Liu, L.; Mou, L.; Vechtomova, O.; and Lin, J. 2019.
\newblock Distilling Task-Specific Knowledge from {BERT} into Simple Neural
  Networks.
\newblock \emph{CoRR} abs/1903.12136.

\bibitem[{Turc et~al.(2019)Turc, Chang, Lee, and Toutanova}]{Well-read}
Turc, I.; Chang, M.; Lee, K.; and Toutanova, K. 2019.
\newblock Well-Read Students Learn Better: The Impact of Student Initialization
  on Knowledge Distillation.
\newblock \emph{CoRR} abs/1908.08962.

\bibitem[{Vaswani et~al.(2017)Vaswani, Shazeer, Parmar, Uszkoreit, Jones,
  Gomez, Kaiser, and Polosukhin}]{Transformer}
Vaswani, A.; Shazeer, N.; Parmar, N.; Uszkoreit, J.; Jones, L.; Gomez, A.~N.;
  Kaiser, L.; and Polosukhin, I. 2017.
\newblock Attention is All you Need.
\newblock In \emph{{NIPS}}, 5998--6008.

\bibitem[{Wang et~al.(2019)Wang, Singh, Michael, Hill, Levy, and Bowman}]{GLUE}
Wang, A.; Singh, A.; Michael, J.; Hill, F.; Levy, O.; and Bowman, S.~R. 2019.
\newblock {GLUE:} {A} Multi-Task Benchmark and Analysis Platform for Natural
  Language Understanding.
\newblock In \emph{{ICLR} (Poster)}. OpenReview.net.

\bibitem[{Wang et~al.(2020)Wang, Wei, Dong, Bao, Yang, and Zhou}]{MiniLM}
Wang, W.; Wei, F.; Dong, L.; Bao, H.; Yang, N.; and Zhou, M. 2020.
\newblock MiniLM: Deep Self-Attention Distillation for Task-Agnostic
  Compression of Pre-Trained Transformers.
\newblock In \emph{NeurIPS}.

\bibitem[{Wang, Wohlwend, and Lei(2019)}]{Ziheng}
Wang, Z.; Wohlwend, J.; and Lei, T. 2019.
\newblock Structured Pruning of Large Language Models.
\newblock \emph{CoRR} abs/1910.04732.

\bibitem[{Xu et~al.(2020)Xu, Zhou, Ge, Wei, and Zhou}]{BERT-of-Theseus}
Xu, C.; Zhou, W.; Ge, T.; Wei, F.; and Zhou, M. 2020.
\newblock BERT-of-Theseus: Compressing {BERT} by Progressive Module Replacing.
\newblock In \emph{{EMNLP} {(1)}}, 7859--7869. Association for Computational
  Linguistics.

\bibitem[{Yang et~al.(2019)Yang, Dai, Yang, Carbonell, Salakhutdinov, and
  Le}]{XLNet}
Yang, Z.; Dai, Z.; Yang, Y.; Carbonell, J.~G.; Salakhutdinov, R.; and Le, Q.~V.
  2019.
\newblock XLNet: Generalized Autoregressive Pretraining for Language
  Understanding.
\newblock In \emph{NeurIPS}, 5754--5764.

\bibitem[{Zhang et~al.(2020)Zhang, Hou, Yin, Shang, Chen, Jiang, and
  Liu}]{TernaryBERT}
Zhang, W.; Hou, L.; Yin, Y.; Shang, L.; Chen, X.; Jiang, X.; and Liu, Q. 2020.
\newblock TernaryBERT: Distillation-aware Ultra-low Bit {BERT}.
\newblock In \emph{{EMNLP} {(1)}}, 509--521. Association for Computational
  Linguistics.

\end{thebibliography}

\end{document}